\documentclass[runningheads,a4paper]{llncs}

\usepackage{amssymb}
\usepackage{graphicx}

\usepackage{amsmath,amssymb}

\usepackage{url}

\begin{document}

\mainmatter  % start of an individual contribution

\title{Provably scale-covariant networks from oriented quasi quadrature measures in cascade
\thanks{In {\em Proc.\ SSVM 2019: Scale Space and Variational Methods
    in Computer Vision\/}, Springer LNCS volume 11603, pages 328--340. The support from the Swedish Research Council 
              (contract 2018-03586) is gratefully acknowledged.}}

\titlerunning{Provably scale-covariant hierarchical networks}

\author{Tony Lindeberg}

\institute{Computational Brain Science Lab\\Division of Computational Science and Technology\\
 KTH Royal Institute of Technology, Stockholm, Sweden}

\maketitle
\begin{abstract}
This article presents a continuous model for hierarchical networks
based on a combination of mathematically derived models of receptive 
fields and biologically inspired computations. 
Based on a functional model of complex cells in terms of an oriented
quasi quadrature combination of first- and second-order directional 
Gaussian derivatives, we couple such primitive computations in cascade
over combinatorial expansions over image orientations.
Scale-space properties of the computational primitives are analysed
and it is shown that the resulting representation allows
for provable scale and rotation covariance. 
A prototype application to texture analysis is developed
and it is demonstrated that a simplified mean-reduced representation of the
resulting QuasiQuadNet leads to promising
experimental results on three texture datasets.
\end{abstract}

\section{Introduction}

The recent progress with deep learning architectures 
has demonstrated that hierarchical feature representations over multiple layers have much higher potential compared to
approaches based on single layers of receptive fields.
A limitation of current deep nets, however, is that they are not truly scale
covariant.
A deep network constructed by repeated application of compact 
$3 \times 3$ or $5 \times 5$ kernels, such as AlexNet \cite{KriSutHin12-NIPS},
VGG-Net \cite{SimZis15-ICLR} or
ResNet \cite{HeZhaRenSun16-CVPR},
implies an implicit assumption of a preferred size in the image
domain as induced by the discretization in terms of local $3 \times 3$ 
or $5 \times 5$ kernels of a fixed size.
Thereby, due to the non-linearities in the deep net, the output from
the network may be qualitatively different
depending on the specific size of the object in the image domain,
as varying because of {\em e.g.\/}\ different distances between the
object and the observer.
To handle this lack of scale covariance, approaches have been developed such as
spatial transformer networks \cite{JadSimZisKav15-NIPS}, using sets of
subnetworks in a multi-scale fashion \cite{CaiFanFerVas16-ECCV} or by combining deep nets with image
pyramids \cite{LinDolGirHeHarBel17-CVPR}.
Since the size normalization performed by a spatial transformer
network is not guaranteed to be truly scale covariant, and since
traditional image pyramids imply a loss of image information
that can be interpreted as corresponding to undersampling,
it is of interest to
develop continuous approaches for deep networks that guarantee true scale covariance or better
approximations thereof.

The subject of this article is to develop a continuous model for capturing
non-linear hierarchical relations between features over multiple
scales in such a way that the resulting feature representation is
provably scale covariant. Building upon axiomatic modelling of visual
receptive fields in terms of Gaussian derivatives and affine
extensions thereof, which can serve as idealized models of simple
cells in the primary visual cortex
\cite{KoeDoo92-PAMI,Lin10-JMIV,Lin13-BICY},
we will propose a functional model for complex cells in terms 
of an oriented quasi quadrature measure.
Then, we will combine such oriented quasi quadrature measures in cascade, building upon
the early idea of Fukushima \cite{Fuk80-BICY} of using 
Hubel and Wiesel's findings regarding
receptive fields in the primary visual cortex \cite{HubWie05-book}
to build a hierarchical neural network from repeated application
of models of simple and complex cells.

We will show how the scale-space properties of the quasi
quadrature primitive in this representation can be theoretically
analyzed and how the resulting hand-crafted network becomes provably scale
and rotation covariant, in such a way that the multi-scale and
multi-orientation network commutes with scaling transformations and
rotations over the spatial image domain. Experimentally, we will investigate a prototype application
to texture classification based on a substantially mean-reduced representation of
the resulting QuasiQuadNet.

\section{The quasi quadrature measure over a 1-D signal}

Consider the scale-space representation $L(x;\; s)$ of a 1-D signal $f(x)$ defined by
convolution with Gaussian kernels $g(x;\; s) =
\exp(-x^2/2s)/\sqrt{2\pi s}$
%\cite{Iij62,Wit83,Koe84,Lin93-Dis,Flo97-book} 
and with 
scale-normalized derivatives according to 
$\partial_{\xi^n} = \partial_{x^n,\gamma-norm}=
s^{n \gamma/2} \, \partial_x^n$ \cite{Lin97-IJCV}.

\paragraph{Quasi quadrature in 1-D.}

Motivated by the fact that the first-order derivatives primarily
respond to the locally odd component of the signal, whereas the
second-order derivatives primarily respond to the locally even
component of a signal, it is natural to aim at a differential feature
detector that combines locally odd and even components in a
complementary manner. By specifically combining the first- and
second-order scale-normalized derivative responses in a 
Euclidean way, we obtain a quasi quadrature measure of the form
\begin{equation}
  \label{eq-quasi-quad-1D}
  {\cal Q}_{x,norm} L 
  = \sqrt{\frac{s \, L_x^2 + C \, s^2 \, L_{xx}^2}{s^{\Gamma}}}
\end{equation}
as a modification
of the quasi quadrature measures previously proposed and studied in \cite{Lin97-IJCV,Lin18-SIIMS},
with the scale normalization parameters $\gamma_1$ and $\gamma_2$
of the first- and second-order derivatives coupled according to
$\gamma_1 = 1 - \Gamma$ and $\gamma_2 = 1 - \Gamma/2$ to enable
scale covariance by adding derivative expressions of different orders
only for the scale-invariant choice of $\gamma = 1$.
This differential entity can be seen as an approximation of the notion
of a quadrature pair of an odd and even filter as more traditionally
formulated based on a Hilbert transform, while confined within
the family of differential expressions based on Gaussian derivatives.

\begin{figure}[hbt]
  \begin{center}
    \begin{tabular}{ccc}
     \hspace{-4mm}
        {\footnotesize\em $g(x;\ s)$\/} 
           & {\footnotesize\em $g_x(x;\ s)$\/} 
           & {\footnotesize\em $g_{xx}(x;\ s)$\/} \\
        \includegraphics[width=0.30\textwidth]{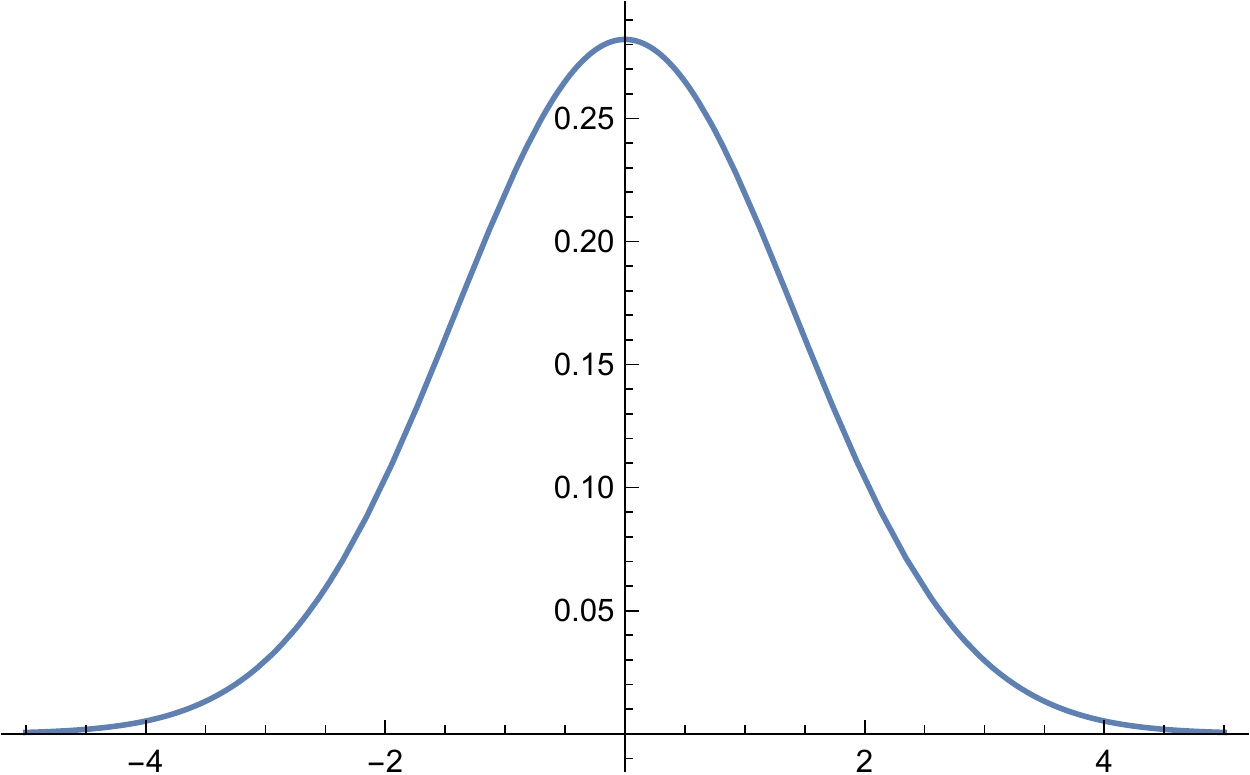}
           & \includegraphics[width=0.30\textwidth]{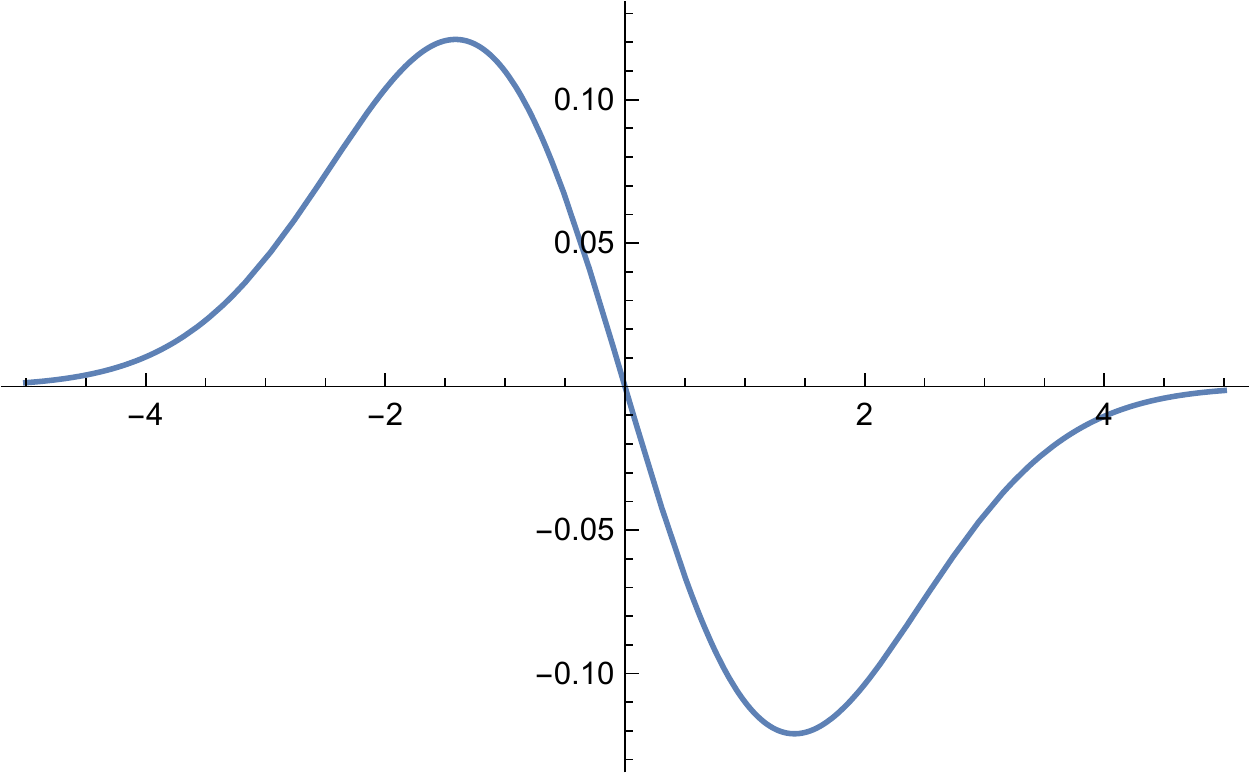} 
           & \includegraphics[width=0.30\textwidth]{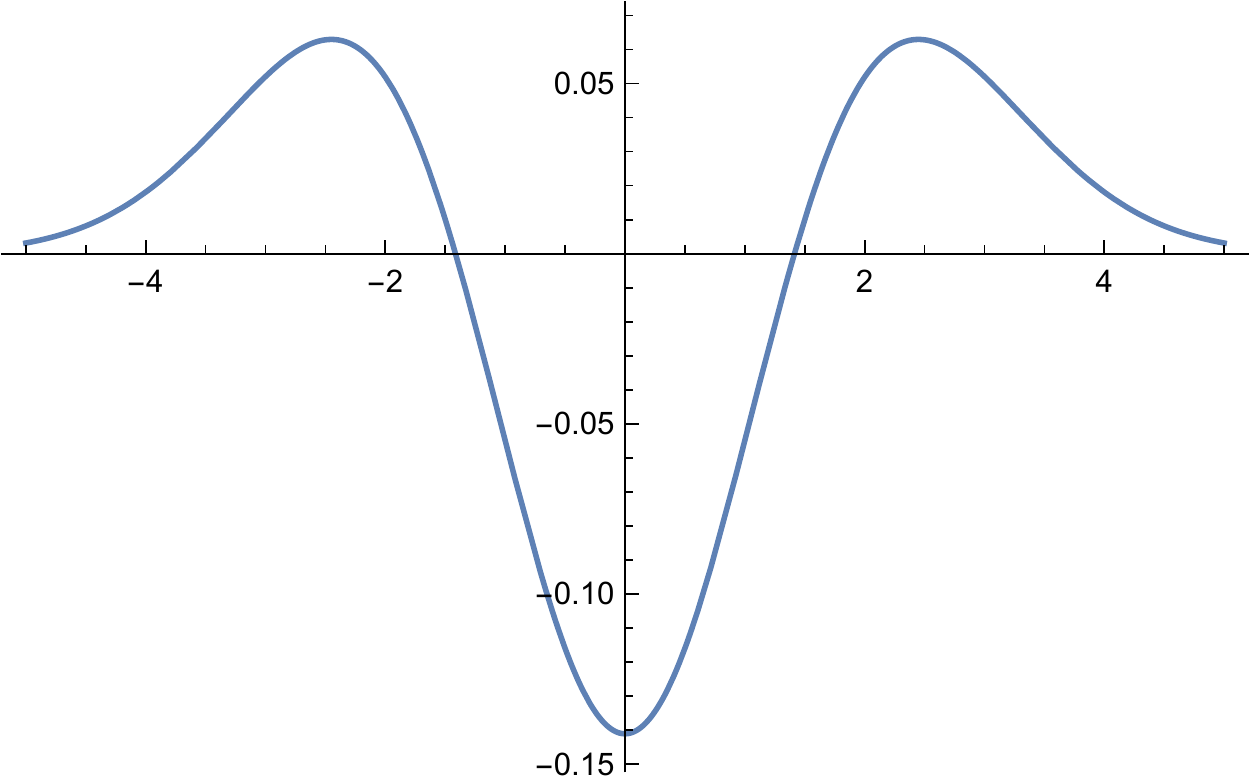} 
           \medskip \\
        {\footnotesize\em ${\cal Q}_{x,norm} L$}
          & {\footnotesize\em ${\cal Q}_{x,norm} L$} 
          & {\footnotesize\em ${\cal Q}_{x,norm} L$} \\
        \includegraphics[width=0.30\textwidth]{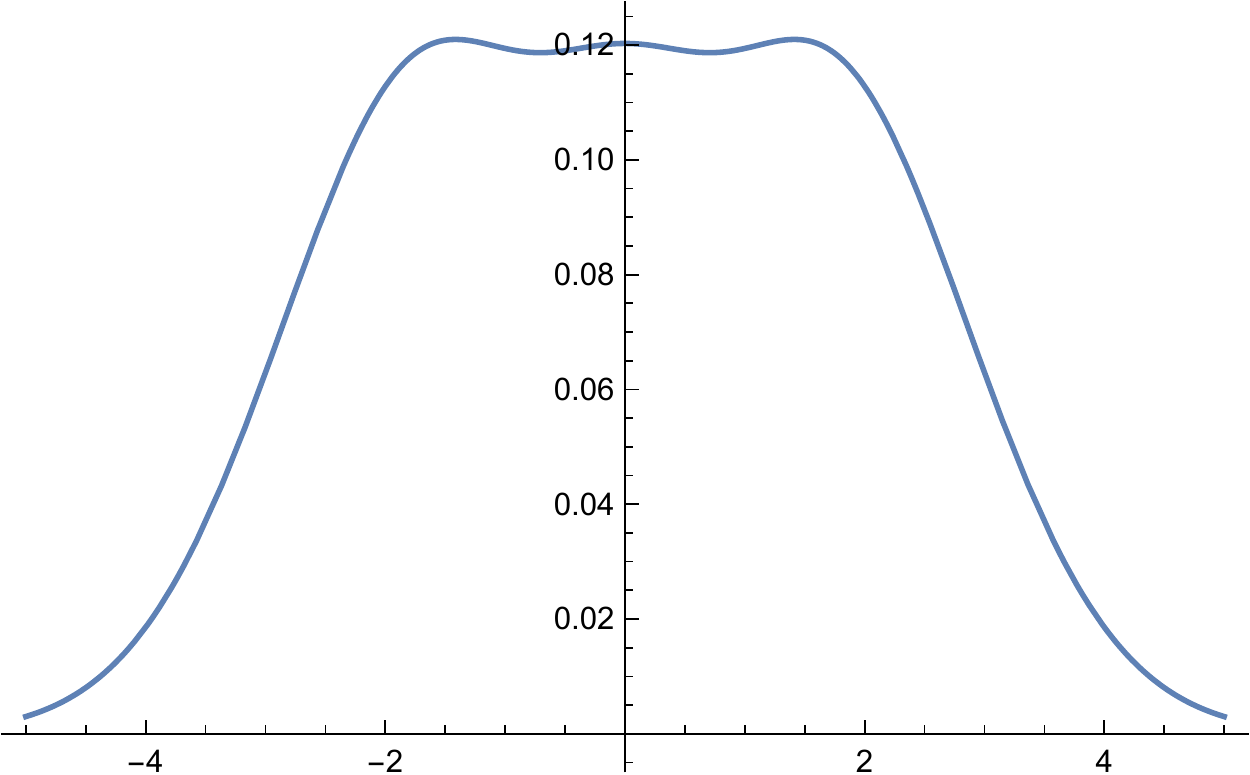} 
          & \includegraphics[width=0.30\textwidth]{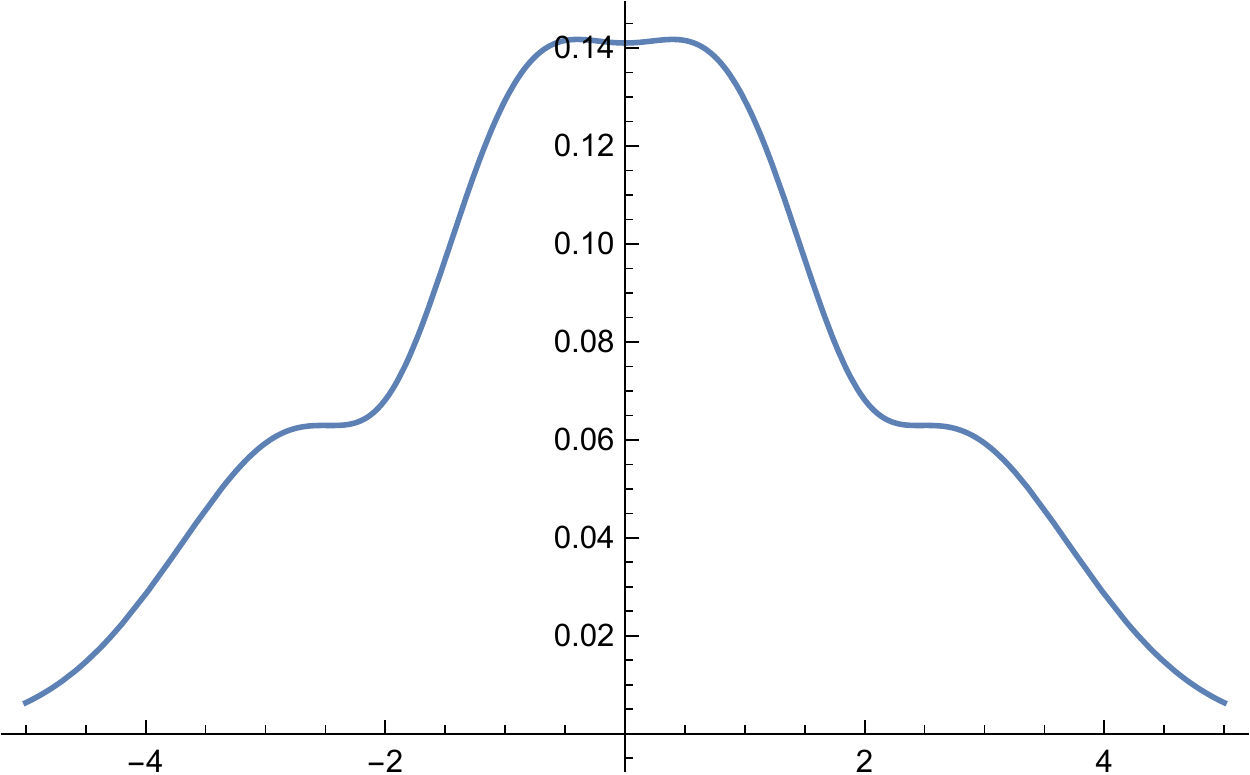} 
          & \includegraphics[width=0.30\textwidth]{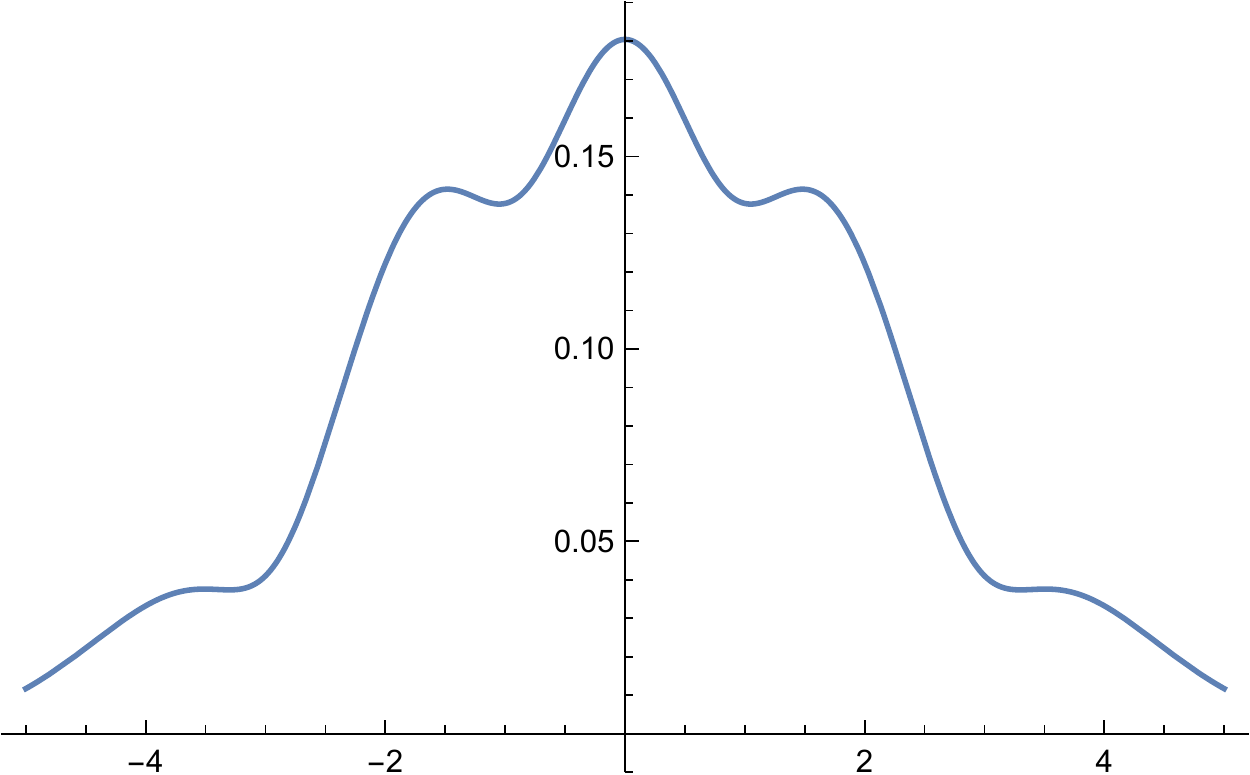} 
          \medskip \\
 \end{tabular} 
  \end{center}
  \vspace{-4mm}
  \caption{1-D Gaussian derivatives up to orders 0, 1 and 2 for $s_0 = 1$ with the
    corresponding 1-D quasi quadrature measures computed from them at
    scale $s = 1$ for $C = 8/11$.
   (Horizontal axis: $x \in [-5, 5]$.)}
  \label{fig-quasiquad-gaussders-1D}
\end{figure}

Figure~\ref{fig-quasiquad-gaussders-1D} shows the result of computing
this quasi quadrature measure for a Gaussian peak as well as its
first- and second-order derivatives. As can be seen, the quasi
quadrature measure is much less sensitive to the position of the peak
compared to {\em e.g.\/}\ the first- or second-order
derivatives. Additionally, the quasi quadrature measure also has some
degree of spatial insensitivity for a first-order
derivative (a local edge model) and a second-order derivative.

\paragraph{Determination of $C$.}

To determine the weighting parameter $C$ between local second-order
and first-order information, let us consider a Gaussian blob $f(x) = g(x;\; s_0)$
with spatial extent given by $s_0$ as input model signal.
By using the semi-group property of the Gaussian kernel
$g(\cdot;\; s_1) * g(\cdot;\; s_2) = g(\cdot;\; s_1 + s_2)$,
the quasi quadrature measure can be computed in closed form
\begin{equation}
   {\cal Q}_{x,norm} L
   = \frac{s^{\frac{1-\Gamma}{2}} e^{-\frac{x^2}{2(s+s_0)}} 
       \sqrt{x^2 (s+s_0)^2 + C s \left(s+s_0-x^2\right)^2+2}}
              {\sqrt{2 \pi}  (s+s_0)^{5/2}}.
\end{equation}
By determining the weighting parameter $C$ such that it minimizes the
overall ripple in the squared quasi quadrature measure for a Gaussian input
\begin{equation}
   \hat{C} = \operatorname{argmin}_{C \geq 0}
                      \int_{x=-\infty}^{\infty} 
                         \left( \partial_x({\cal Q}^2_{x,norm} L) \right)^2 \, dx,
\end{equation}
we obtain
\begin{equation}
  \hat{C} = \frac{4 (s+s_0)}{11 s},
\end{equation}
which in the special case of choosing $s = s_0$ corresponds to $C = 8/11 \approx 0.727$.
This value is very close to the value $C = 1/\sqrt{2} \approx 0.707$
derived from an equal contribution condition in
\cite[Eq.~(27)]{Lin18-SIIMS} for the special case of choosing $\Gamma = 0$.

\paragraph{Scale selection properties.}

To analyze the scale selection properties of the quasi quadrature
measure, let us consider the result of using Gaussian derivatives of
orders 0, 1 and 2 as input signals, {\em i.e.\/},
$f(x) = g_{x^n}(x;\; s_0)$ for $n \in \{ 0, 1, 2 \}$.

For the zero-order Gaussian kernel, the scale-normalized quasi
quadrature measure at the origin is given by
\begin{equation}
  \left. {\cal Q}_{x,norm} L \right|_{x=0,n=0}
  = \frac{\sqrt{C} s^{1-\Gamma/2}}{2 \pi (s+s_0)^2}.
\end{equation}
For the first-order Gaussian derivative kernel, the scale-normalized quasi
quadrature measure at the origin is 
\begin{equation}
  \left. {\cal Q}_{x,norm} L \right|_{x=0,n=1}
  = \frac{s_0^{1/2} s^{(1-\Gamma)/2}}{2 \pi (s+s_0)^2}
\end{equation}
whereas for the second-order Gaussian derivative kernel,
the scale-normalized quasi quadrature measure at the origin is
\begin{equation}
  \left. {\cal Q}_{x,norm} L \right|_{x=0,n=2}
  = \frac{3 \sqrt{C} s_0 s^{1-\Gamma/2}}{2 \pi  (s+s_0)^3}.
\end{equation}
By differentiating these expressions with respect to scale, we find
that for a zero-order Gaussian kernel the maximum response over scale
is assumed at
\begin{equation}
   \left. \hat{s} \right|_{n=0} 
   = \frac{s_0 \, (2 -\Gamma)}{2+\Gamma},
\end{equation}
whereas for first- and second-order derivatives, respectively, the
maximum response over scale is assumed at
\begin{align}
  \begin{split}
     \left. \hat{s} \right|_{n=1} 
     = \frac{s_0 \; (1 -\Gamma)}{3+\Gamma}, \quad\quad
     \left. \hat{s} \right|_{n=2} 
     = \frac{s_0 \, (2 - \Gamma)}{4+\Gamma}.
  \end{split}
\end{align}
In the special case of choosing $\Gamma = 0$, these scale estimates
correspond to 
\begin{equation}
  %==>> Hand patched equation numbers
  \tag{10a-c}
  %\label{eq-scsel-quasiquad-012-ders}
   \left. \hat{s} \right|_{n=0} 
     = s_0, \quad\quad
    \left. \hat{s} \right|_{n=1} 
     = \frac{s_0}{3}, \quad\quad
    \left. \hat{s} \right|_{n=2} 
     = \frac{s_0}{2}.
  \addtocounter{equation}{+1}
\end{equation}
Thus, for a Gaussian input signal, the selected scale level will for
the most scale-invariant choice of using $\Gamma = 0$ reflect the
spatial extent $\hat{s} = s_0$ of the blob, whereas if we would like
the scale estimate to reflect the scale parameter of first- and
second-order derivatives, we would have to choose $\Gamma = -1$.
An alternative motivation for using finer scale levels for the
Gaussian derivative kernels is to regard the positive and negative lobes
of the Gaussian derivative kernels as substructures of a more complex
signal, which would then warrant the use of finer scale levels to
reflect the substructures of the signal 
%==>> Hand patched equation numbers to %\label{eq-scsel-quasiquad-012-ders}
((10b) and (10c)).

\section{Oriented quasi quadrature modelling of complex cells}

In this section, we will consider an extension of the 1-D quasi
quadrature measure (\ref{eq-quasi-quad-1D}) into an oriented quasi quadrature measure of the form
\begin{equation}
  \label{eq-quasi-quad-dir}
  {\cal Q}_{\varphi,norm} L 
  = \sqrt{\frac{\lambda_{\varphi} \, L_{\varphi}^2 + C \, \lambda_{\varphi}^2 \, L_{\varphi\varphi}^2}{s^{\Gamma}}},
\end{equation}
where $L_{\varphi}$ and $L_{\varphi\varphi}$ denote directional
derivatives of an affine Gaussian scale-space representation \cite[ch.~15]{Lin93-Dis} of the form 
    $L_{\varphi} 
     = \cos \varphi \, L_{x_1} + \sin \varphi \, L_{x_2}$ and
     $L_{\varphi\varphi} 
     = \cos^2 \varphi \, L_{x_1x_1} + 2 \cos \varphi \, \sin \varphi \, L_{x_1x_2} + \sin^2 \varphi \, L_{x_2x_2}$,
and with
$\lambda_{\varphi}$ denoting the variance of the affine
Gaussian kernel (with $x = (x_1, x_2)^T$)
\begin{equation}
  \label{eq-aff-gauss-2D}
   g(x;\; s, \Sigma)  = \frac{1}{2 \pi s \sqrt{\det\Sigma}} e^{-x^T \Sigma^{-1} x/2s} 
  \end{equation}
in direction $\varphi$, preferably with the orientation $\varphi$
aligned with the direction $\alpha$ of either of the eigenvectors of the
composed spatial covariance matrix $s \, \Sigma$, with 
\begin{align}
  \begin{split}
  \label{eq-aff-cov-mat-2D}
  \Sigma & =
  \frac{1}{\max(\lambda_1, \lambda_2)}
  \left(
    \begin{array}{ccc}
      \lambda_1 \cos^2 \alpha + \lambda_2 \sin^2 \alpha \quad
        &  (\lambda_1 - \lambda_2) \cos \alpha \, \sin \alpha 
        \\
      (\lambda_1 - \lambda_2) \cos \alpha \, \sin \alpha \quad
        & \lambda_1 \sin^2 \alpha + \lambda_2 \cos^2 \alpha
    \end{array}
  \right)
  \end{split}
\end{align}
normalized such that the main eigenvalue is equal to one.

\begin{figure}[hbtp]
   \vspace{-4mm}
    \begin{center}
     \begin{tabular}{ccc}
        & & {\small $\partial_{\varphi}g(x, y;\; \Sigma)$} \\
       \includegraphics[height=0.17\textheight]{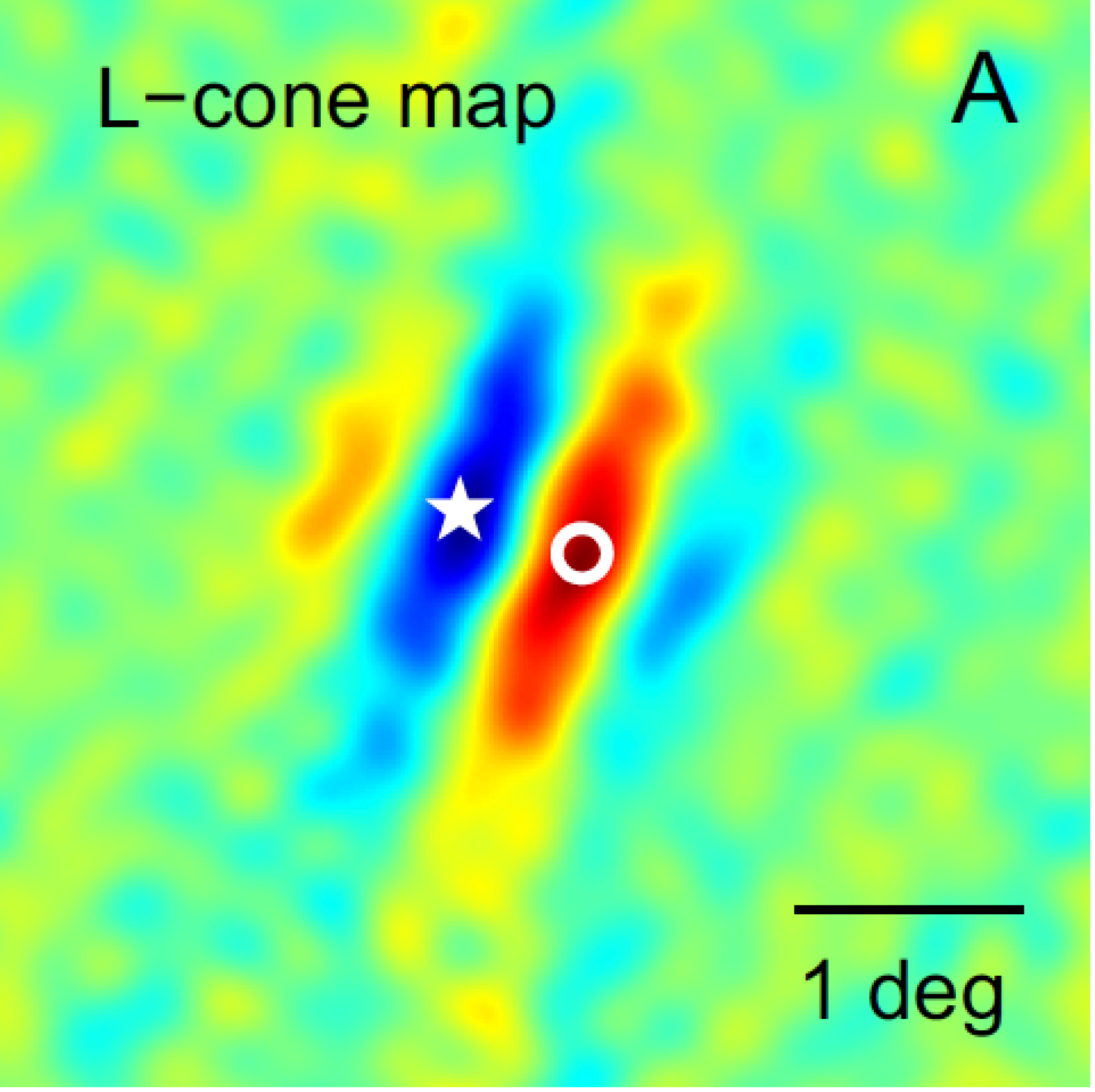}
       & \includegraphics[height=0.17\textheight]{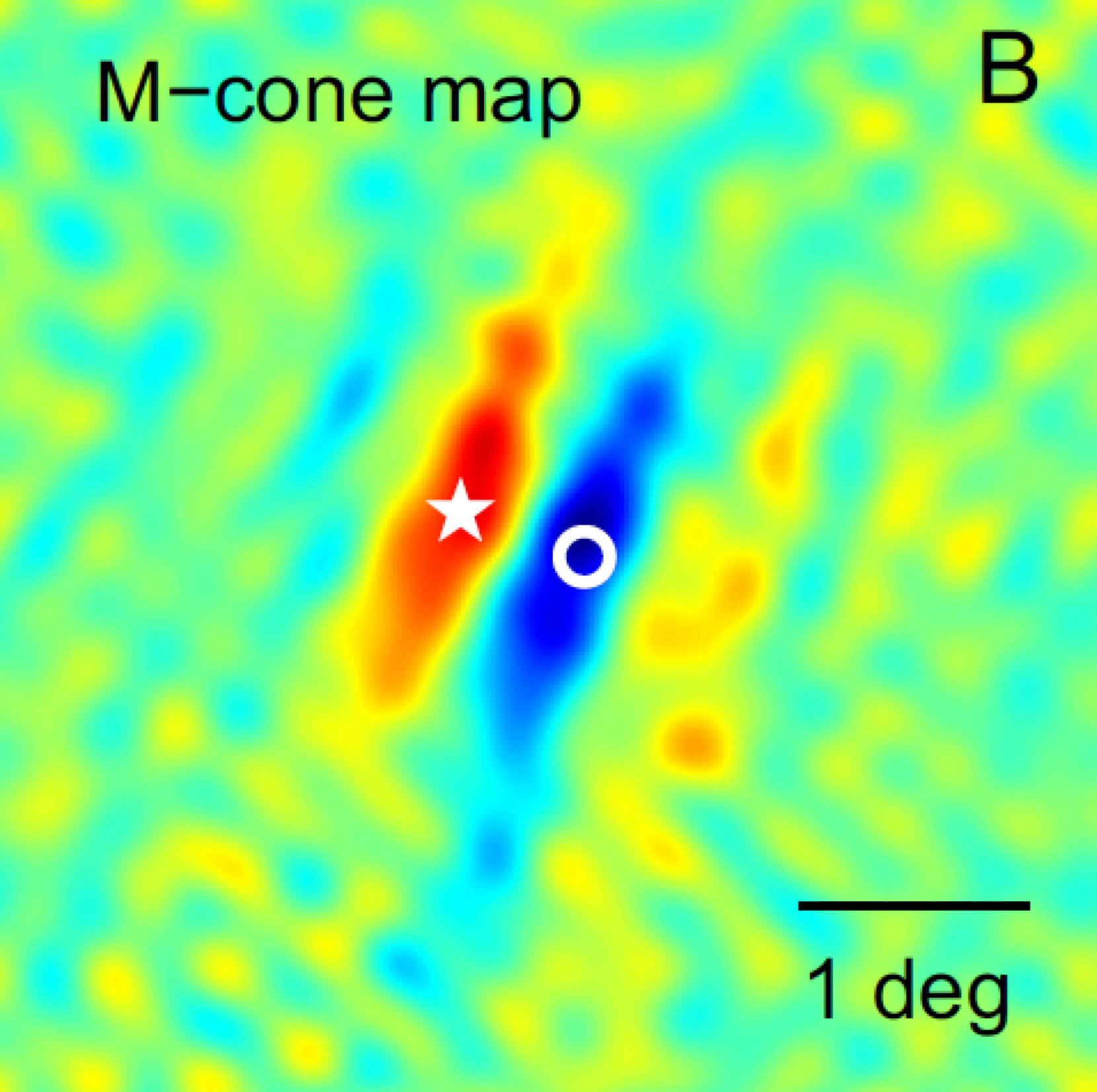}
       & \includegraphics[height=0.17\textheight]{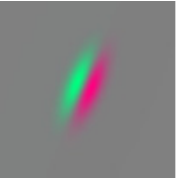}
     \end{tabular}
   \end{center}
   \vspace{-2mm}
  \caption{Example of a colour-opponent receptive field profile 
           for a double-opponent simple cell in the primary visual
           cortex (V1) as measured by Johnson {\em et al.\/}\
           \cite{JohHawSha08-JNeuroSci}.
          (left) Responses to L-cones corresponding to long wavelength
          red cones, with positive weights
          represented by red and negative weights by blue. 
          (middle) Responses to M-cones corresponding to medium wavelength
          green cones, with positive weights
          represented by red and negative weights by blue. 
          (right) Idealized model of the receptive field
          from a first-order directional derivative of an affine
          Gaussian kernel $\partial_{\varphi}g(x, y;\; \Sigma)$ 
          according to (\ref{eq-spat-RF-model}) for $\sigma_1 = \sqrt{\lambda_1} = 0.6$,
         $\sigma_2 = \sqrt{\lambda_2} = 0.2$ in units of
           degrees of visual angle, $\alpha = 157~\mbox{degrees}$ and with positive
           weights for the red-green colour-opponent channel
           $U = R-G$ with positive values represented by red and
           negative values by green.}
  \label{fig-simple-cell-aff-gauss-model-col-opp}
  \vspace{-4mm}
\end{figure}

\paragraph{Affine Gaussian derivative model for linear receptive
  fields.}

According to the normative theory for visual receptive fields in
Lindeberg \cite{Lin10-JMIV,Lin13-BICY},
directional derivatives of affine Gaussian kernels
constitute a canonical model for visual receptive fields over a 2-D
spatial domain.
Specifically, it was proposed that simple cells in the primary visual
cortex (V1) can be modelled by directional derivatives of affine Gaussian
kernels, termed {\em affine Gaussian derivatives\/}, of the form
\begin{equation}
  \label{eq-spat-RF-model}
   T_{{\varphi}^{m}}(x_1, x_2;\; s, \Sigma)  
  = \partial_{\varphi}^{m} 
      \left( g(x_1, x_2;\; s, \Sigma) \right).
\end{equation}
Figure~\ref{fig-simple-cell-aff-gauss-model-col-opp} shows an example of the
spatial dependency of a colour-opponent simple cell that can be well modelled by a
first-order affine Gaussian derivative over an R-G colour-opponent
channel over image intensities. 
Corresponding modelling results for non-chromatic receptive fields can be found in \cite{Lin10-JMIV,Lin13-BICY}.

\begin{figure}[hbtp]
  %\vspace{-5mm}
   \begin{center}
     \begin{tabular}{c}
       \includegraphics[width=0.60\textwidth]{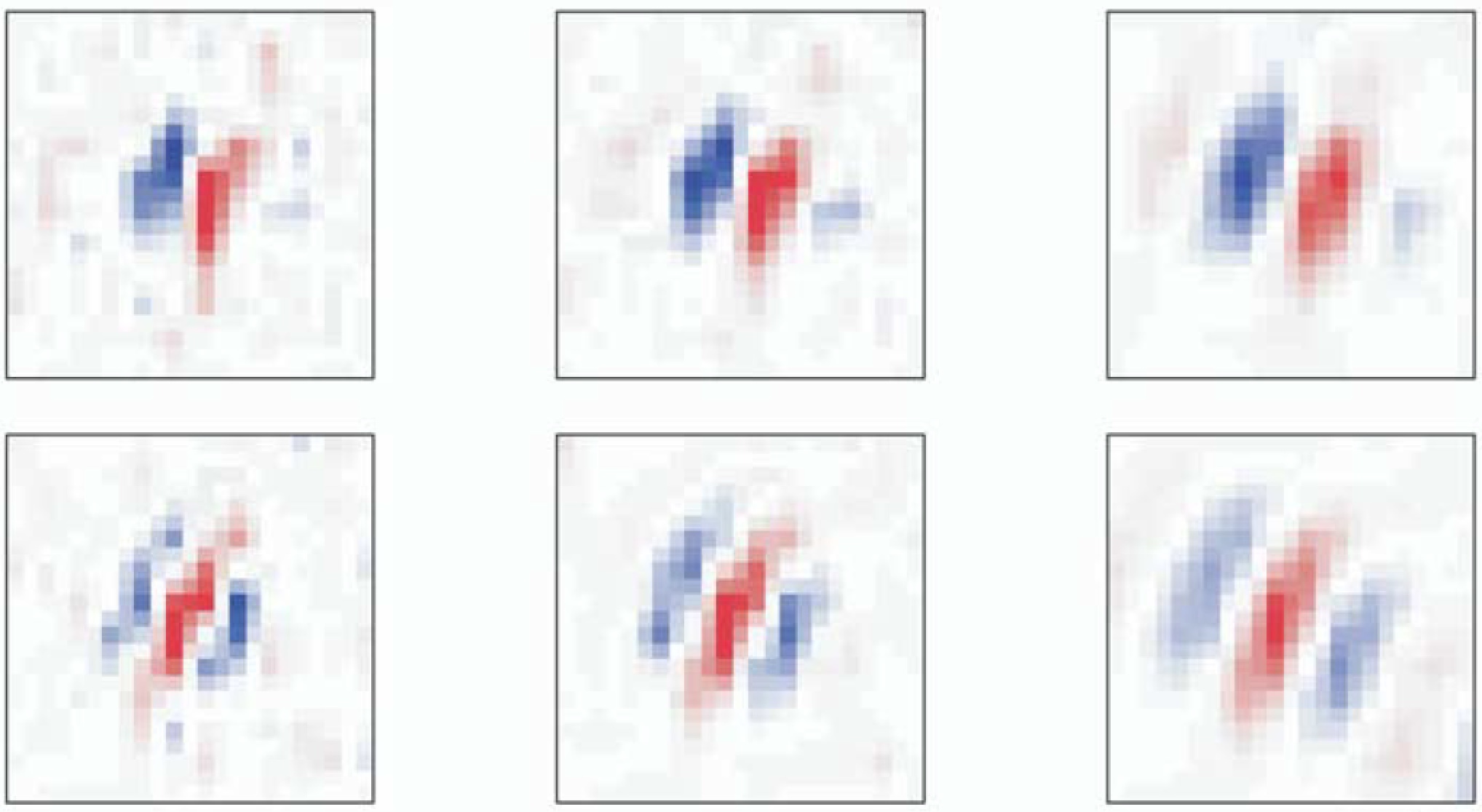}
     \end{tabular}
   \end{center}
   \vspace{-3mm}
   \caption{Significant eigenvectors of a complex cell in the cat
     primary visual cortex, as
     determined by Touryan {\em et al.\/}\
\cite{TouFelDan05-Neuron} from the response properties of the cell to a set of natural image
   stimuli, using a spike-triggered covariance method (STC) that
   computes the eigenvalues and the eigenvectors of a second-order
   Wiener kernel using three different parameter settings (cutoff
   frequencies) in the system
   identification method (from left to right).
   Qualitatively, these kernel shapes agree well with the shapes of
   first- and second-order affine Gaussian derivatives.}
  \label{fig-Touryan-eigenvectors-complex-cell}
  \vspace{-4mm}
\end{figure}

\paragraph{Affine quasi quadrature modelling of complex cells.}

Figure~\ref{fig-Touryan-eigenvectors-complex-cell} shows functional properties of a complex cell as
determined from its response properties to natural images, 
using a spike-triggered covariance method (STC),
which computes the eigenvalues and the eigenvectors of
a second-order Wiener kernel (Touryan {\em et al.\/}\
\cite{TouFelDan05-Neuron}).
As can be seen from this figure, the shapes of the eigenvectors 
determined from the non-linear Wiener kernel model of the complex cell
do qualitatively agree very well with the shapes of corresponding
affine Gaussian derivative kernels of orders 1 and 2.
Motivated by this property and theoretical and experimental
motivations for modelling receptive field profiles of simple cells by
affine Gaussian derivatives, we propose to model complex cells by a
possibly post-smoothed (spatially pooled) oriented quasi quadrature measure of the
form (\ref{eq-quasi-quad-dir})
\begin{equation}
  \label{eq-quasi-quad-dir-smooth}
  (\overline{\cal Q}_{\varphi,norm} L)(\cdot;\; s_{loc}, s_{int}, \Sigma_{\varphi})
  = \sqrt{g(\cdot;\; s_{int}, \Sigma_{\varphi}) 
               * ({\cal Q}^2_{\varphi,norm} L)(\cdot;\; s_{loc}, \Sigma_{\varphi})} 
\end{equation}
where $s_{loc} \,\Sigma_{\varphi}$ represents an affine covariance matrix 
in direction $\varphi$ for computing directional
derivatives and $s_{int} \, \Sigma_{\varphi}$ represents an affine covariance
matrix in the same direction for integrating pointwise affine quasi quadrature
measures over a region in image space.

The pointwise affine quasi quadrature measure
$({\cal Q}_{\varphi,norm} L)(\cdot;\; s_{loc}, \Sigma_{\varphi})$
can be seen as a Gaussian derivative based analogue
of the energy model for complex cells as proposed by 
Adelson and Bergen \cite{AdeBer85-JOSA} and Heeger \cite{Hee92-VisNeuroSci}.
It is closely related to a proposal by 
Koenderink and van Doorn \cite{KoeDoo90-BC} of summing up the squares of
first- and second-order derivative responses and
nicely compatible with results by 
De~Valois {\em et al.\/}\ \cite{ValCotMahElfWil00-VR}, who showed that
first- and second-order receptive fields typically occur in pairs that
can be modelled as approximate Hilbert pairs. 

The addition of a complementary post-smoothing stage as
determined by the affine Gaussian weighting function $g(\cdot;\; s_{int}, \Sigma_{\varphi})$
is closely related to recent results by West{\"o} and May \cite{WesMay18-JNeuroPhys},
who have shown that complex cells are better modelled as a combination
of two spatial integration steps. 

By choosing these spatial smoothing
and weighting functions as affine Gaussian kernels, we ensure an affine covariant model of the
complex cells, to enable the computation of affine invariants at higher
levels in the visual hierarchy.

The use of multiple affine receptive fields over different shapes of the affine
covariance matrices $\Sigma_{\varphi,loc}$ and $\Sigma_{\varphi,int}$ can
be motivated by results by Goris {\em et al.\/}\
\cite{GorSimMov15-Neuron}, who show that there is a large variability
in the orientation selectivity of simple and complex cells.
%
% ==>> (see Figure~\ref{fig-DeValois-orient-select-stat}). 
%
With respect
to this model, this means that we can think of affine covariance
matrices of different eccentricity as being present from isotropic to
highly eccentric. By considering the full family of positive definite
affine covariance matrices, we obtain a fully affine covariant
image representation able to handle local linearizations of the perspective
mapping for all possible views of any smooth local surface patch.

\section{Hierarchies of oriented quasi quadrature measures}

Let us in this first study disregard the variability due to different shapes
of the affine receptive fields for different eccentricities and assume
that $\Sigma = I$.
This restriction enables covariance to scaling transformations and rotations,
whereas a full treatment of affine quasi quadrature measures over all
positive definite covariance matrices
would have the potential to enable full affine covariance.

An approach that we shall pursue is to build feature
hierarchies by coupling oriented quasi quadrature measures
(\ref{eq-quasi-quad-dir}) or (\ref{eq-quasi-quad-dir-smooth}) in
cascade 
\begin{align}
  \begin{split}
    \label{eq-hier-quasi-quad-line1}
     & F_1(x, \varphi_1) = ({\cal Q}_{\varphi_1,norm} \, L)(x) 
  \end{split}\\
  \begin{split}
    \label{eq-hier-quasi-quad-line2}
      & F_k(x, \varphi_1, ..., \varphi_{k-1}, \varphi_k) = 
      ({\cal Q}_{\varphi_k,norm} \, F_{k-1})(x, \varphi_1, ..., \varphi_{k-1}),
   \end{split}
\end{align}
where we have suppressed the notation for the scale
levels assumed to be distributed such that the scale
parameter at level $k$ is $s_k = s_0 \, r^{2(k-1)}$ for some 
$r > 1$, {\em e.g.}, $r = 2$. 
Assuming that the initial scale-space representation $L$ is
computed at scale $s_0$, such a network can in turn be initiated for
different values of $s_0$, also distributed according to a geometric distribution.

This construction builds upon an early proposal by Fukushima \cite{Fuk80-BICY}
of building a hierarchical neural network from repeated application
of models of simple and complex cells \cite{HubWie05-book},
which has later been explored in a hand-crafted network based on
Gabor functions by Serre {\em et al.\/}\ \cite{SerWolBilRiePog07-PAMI}
and in the scattering convolution networks by Bruno and Mallat
\cite{BruMal13-PAMI}. This idea is also consistent with a proposal by
Yamins and DiCarlo \cite{YamDiC16-NatNeuroSci} of using repeated application 
of a single hierarchical convolution layer for explaining the
computations in the mammalian cortex.
With this construction, we obtain a way to
define continuous networks that express a corresponding hierarchical
architecture based on Gaussian derivative based models of simple and
complex cells within the scale-space framework.

Each new layer in this model implies an expansion of combinations
of angles over the different layers in the hierarchy. For example, if
we in a discrete implementation discretize the angles $\varphi \in [0, \pi[$ into $M$ discrete
spatial orientations, we will then obtain $M^k$ different features at
level $k$ in the hierarchy. To keep the complexity down at higher
levels, we will for $k \geq K$ in a corresponding way as done by
Hadji and Wildes \cite{HadWil17-ICCV} introduce a pooling stage 
over orientations 
\begin{equation}
  ({\cal P}_k F_{k})(x, \varphi_1, ..., \varphi_{K-1}) 
   = \sum_{\varphi_k} F_k(x, \varphi_1, ..., \varphi_{K-1}, \varphi_k),
\end{equation}
and instead define the next successive layer as
\begin{equation}
     F_k(x, \varphi_1, ..., \varphi_{k-2}, \varphi_{K-1}, \varphi_k) = 
     ({\cal Q}_{\varphi_k,norm} \, {\cal P}_{k-1} F_{k-1})(x, \varphi_1, ..., \varphi_{K-1})
\end{equation}
to limit the number of features at any level to maximally
$M^{K-1}$. The proposed hierarchical feature representation 
is termed QuasiQuadNet.

\paragraph{Scale covariance.}

A theoretically attractive property of this family of networks is that
the networks are provably scale covariant. Given two images $f$ and $f'$ that
are related by a uniform scaling transformation $f(x) = f'(S x)$ for
some $S > 0$, their corresponding scale-space representations $L$ and
$L'$ will be equal $L'(x';\; s') = L(x;\; s)$ and so will the
scale-normalized derivatives 
$s'^{n/2} \, L'_{{x_i'}^n}(x';\; s') = s^{n/2} \, L_{x_i^n}(x;\; s)$
based on $\gamma = 1$ if the spatial
positions are related according to $x' = S x$  
and the scale levels according to $s' = S^2 s$ 
\cite[Eqns.~(16) and (20)]{Lin97-IJCV}. 
This implies that if the initial scale levels
$s_0$ and $s_0'$ underlying the construction in 
(\ref{eq-hier-quasi-quad-line1}) and (\ref{eq-hier-quasi-quad-line2})
are related according to $s_0' = S^2 s_0$,
then the
first layers of the feature hierarchy will be related according to 
$F_1'(x', \varphi_1) = S^{-\Gamma} \, F_1(x, \varphi_1)$
\cite[Eqns.~(55) and (63)]{Lin18-SIIMS}. Higher
layers in the feature hierarchy are in turn related according to
\begin{equation}
   F_k'(x', \varphi_1, ..., \varphi_{k-1}, \varphi_k) 
   = S^{-k \Gamma} \, F_k(x, \varphi_1, ..., \varphi_{k-1}, \varphi_k)  
\end{equation}
and are specifically equal if $\Gamma = 0$. This means that it will be
possible to perfectly match such hierarchical representations
under uniform scaling transformations.

\paragraph{Rotation covariance.} Under a rotation of image space
by an angle $\alpha$, $f'(x') = f(x)$ for $x'= R_{\alpha} x$, the
corresponding feature hierarchies
are in turn equal if the orientation angles are related according to $\varphi'_i = \varphi_i + \alpha$ ($i = 1..k$)
\begin{equation}
   F_k'(x', \varphi'_1, ..., \varphi'_{k-1}, \varphi'_k) 
   = F_k(x, \varphi_1, ..., \varphi_{k-1}, \varphi_k).
\end{equation}

\section{Application to texture analysis}

In the following, we will use a substantially reduced version of the
proposed quasi quadrature network for building an application to texture
analysis.

If we make the assumption that a spatial texture should obey certain
stationarity properties over image space, we may regard it as reasonable
to construct texture descriptors by accumulating statistics of
feature responses over the image domain, in terms of {\em e.g\/} mean values or histograms.
Inspired by the way the SURF descriptor \cite{BayEssTuyGoo08-CVIU} accumulates mean values and
mean absolute values of derivative responses and the way Bruno and Mallat
\cite{BruMal13-PAMI} and Hadji and
Wildes \cite{HadWil17-ICCV} compute mean values of their hierarchical
feature representations, we will initially explore reducing
the QuasiQuadNet to just the mean values over the image domain of the following 5 features
\begin{equation}
\{ \partial_{\varphi} F_{k}, |\partial_{\varphi}
F_{k}|, \partial_{\varphi\varphi} F_{k}, |\partial_{\varphi\varphi}
F_{k}|, {\cal Q}_{\varphi} F_{k} \}.
\end{equation}
These types of features are computed for all layers in the feature
hierarchy (with $F_0 = L$), which leads to a 4000-D descriptor based on
$M = 8$ uniformly distributed orientations in $[0, \pi[$, 4
layers in the hierarchy delimited in complexity by directional pooling for $K = 3$
with 4 initial scale levels
$\sigma_0 = \sqrt{s_0} \in \{ 1, 2, 4, 8 \}$.

\begin{table}
  \begin{center}
  \begin{tabular}{lccc}
      \hline
      & KTH-TIPS2b & CUReT & UMD\\
      \hline
      FV-VGGVD \cite{CimMajVed15-CVPR} (SVM) & 88.2 & 99.0 & 99.9 \\ % 287.1
      FV-VGGM \cite{CimMajVed15-CVPR} (SVM) & 79.9 & 98.7 & 99.9 \\ % 278.5
      MRELBP \cite{LiuLaoFieGuoWanPie16-TIP}  (SVM) & 77.9 & 99.0 & 99.4 \\ % 276.3 
      FV-AlexNet \cite{CimMajVed15-CVPR}  (SVM) & 77.9 & 98.4 & 99.7 \\ % 276.0
      {\em mean-reduced QuasiQuadNet LUV (SVM)}  & 78.3 & 98.6 \\ % 274.0 
      {\em mean-reduced QuasiQuadNet grey (SVM)}  & 75.3 & 98.3 & 97.1 \\ % 270.7 
      ScatNet \cite{BruMal13-PAMI} (PCA) & 68.9 & 99.7 & 98.4 \\ %267.0 
      MRELBP \cite{LiuLaoFieGuoWanPie16-TIP} & 69.0 & 97.1 & 98.7 \\ % 264.8 
      BRINT \cite{LiuLonFieLaoZha14-TIP}  & 66.7 & 97.0 & 97.4 \\ % 261.5 
      MDLBP \cite{SchDos12-ICPR} & 66.5 & 96.9 & 97.3 \\ %260.7 
      {\em mean-reduced QuasiQuadNet LUV (NNC)}  & 72.1 & 94.9 \\ % 260.3 
      {\em mean-reduced QuasiQuadNet grey (NNC)}  & 70.2 & 93.0 & 93.3 \\ % 256.5
      LBP \cite{OjaPieMae02-PAMI} & 62.7 & 97.0 & 96.2 \\ % 255.9
      ScatNet \cite{BruMal13-PAMI} (NNC) & 63.7 & 95.5 & 93.4 \\ % 252.6
      PCANet \cite{ChaJiaGaoLuZenMa15-TIP} (NNC) & 59.4 & 92.0 & 90.5 \\ % 241.9
      RandNet \cite{ChaJiaGaoLuZenMa15-TIP} (NNC) & 56.9 & 90.9 & 90.9 \\ % 238.7
      \hline
   \end{tabular}
   \end{center}
   \caption{Performance results of the mean-reduced QuasiQuadNet in
     comparison with a selection of among the better methods in the
     extensive performance evaluation by Liu {\em et al.\/}\ \cite{LiuFieGuoWanPei17-PattRecogn}
 (our results in slanted font).}
   \label{tab-perf-KTH-TIPS2b}
   \vspace{-4mm}
\end{table}

The second column in Table~\ref{tab-perf-KTH-TIPS2b} shows the result of applying this approach to the
KTH-TIPS2b dataset \cite{KTH-TIPS2} for texture classification,
consisting of 11 classes (``aluminum foil'', ``cork'',
``wool'', ``lettuce leaf'', ``corduroy'', ``linen'', ``cotton'',
``brown bread'', ``white bread'', ``wood'' and ``cracker") with 4
physical samples from each class and photos of each sample taken from
9 distances leading to 9 relative scales labelled ``2'', \dots, ``10''
over a factor of 4 in scaling transformations and additionally 12 different pose and illumination conditions for each
scale, leading to a total number of $11 \times 4 \times 9 \times 12 = 4752$ images. The regular benchmark setup
implies that the images from 3 samples in each class are used for
training and the remaining sample in each class is used for testing over 4 permutations. Since
several of the samples from the same class are quite different from each other in appearance, this implies
a non-trivial benchmark which has not yet been saturated.

When using nearest-neighbour classification on the mean-reduced grey-level
descriptor, we get 70.2~\% accuracy, and 72.1~\% accuracy when 
computing corresponding features from the LUV channels of a
colour-opponent representation. When using SVM classification, the
accuracy becomes 75.3~\% and 78.3~\%, respectively. Comparing with the results
of an extensive set of other methods in Liu {\em et al.\/}
\cite{LiuFieGuoWanPei17-PattRecogn}, out of which a selection of
the better results are listed in Table~\ref{tab-perf-KTH-TIPS2b},
the results of the mean-reduced QuasiQuadNet are better than
classical texture classification methods such as locally binary patterns (LBP) \cite{OjaPieMae02-PAMI}, 
binary rotation invariant noise tolerant texture descriptors
\cite{LiuLonFieLaoZha14-TIP}
 and multi-dimensional local binary patterns (MDLBP) \cite{SchDos12-ICPR}  
and also better than
other handcrafted networks,
such as ScatNet \cite{BruMal13-PAMI}, PCANet \cite{ChaJiaGaoLuZenMa15-TIP} and RandNet \cite{ChaJiaGaoLuZenMa15-TIP}.
The performance of the mean-reduced QuasiQuadNet descriptor does,
however, not reach the performance of applying SVM classification to
Fischer vectors of the filter output in learned convolutional networks 
(FV-VGGVD, FV-VGGM \cite{CimMajVed15-CVPR}).

By instead performing the training on every second scale in the
dataset (scales 2, 4, 6, 8, 10) and the testing on the other scales
(3, 5, 7, 9), such that the benchmark does not primarily test the
generalization properties between the different very few samples in each class,
the classification performance is 98.8~\% for the
grey-level descriptor and 99.6~\% for the LUV descriptor.

The third and fourth columns in Table~\ref{tab-perf-KTH-TIPS2b} show corresponding results of texture
classification on the CUReT \cite{VarZis09-PAMI} and UMD
\cite{XuYanLinJi10-CVPR} texture datasets, with random equally
sized partitionings of the images into training and testing data.
Also for these datasets, the performance of the 
mean-reduced descriptor is reasonable compared to other methods.

\section{Summary and discussion}

We have presented a theory for defining hand-crafted hierarchical
networks by applying quasi quadrature responses of first- and
second-order directional Gaussian derivatives in cascade.
The purpose behind this study has been to investigate if we could
start building a bridge between the well-founded theory of scale-space
representation and the recent empirical developments in deep learning, while
at the same time being inspired by biological vision.
The present work is intended as an initial work in this direction,
where we propose the family of quasi quadrature networks as a new
baseline for hand-crafted networks with associated provable covariance properties
under scaling and rotation transformations.

By early experiments with a substantially mean-reduced representation of
the resulting QuasiQuadNet, we have demonstrated that it is possible 
to get quite promising performance on texture classification,
and comparable or better than other hand-crafted networks,
although not reaching the performance of 
%applying more refined statistical classification methods on 
learned CNNs.
By inspection of the full non-reduced feature maps, which could not be
shown here because of the space limitations, we have also observed
that some representations in higher layers may respond to irregularities in
regular textures (defect detection) or corners or end-stoppings in
regular scenes.

Concerning extensions of the approach, we propose to:
(i)~complement the computation of quasi quadrature responses by divisive
normalization \cite{CarHee12-NatureRevNeuroSci} to enforce a
competition between multiple feature responses,
(ii)~explore the spatial relationships in the full feature maps that
are suppressed in the mean-reduced representation and
(iii)~incorporate learning mechanisms. 

\vspace{-4mm}

$\,$ 

{\footnotesize
\bibliographystyle{splncs}
\bibliography{bib/defs,bib/tlshort,bib/tlmac}}

\end{document}